\documentstyle[cupbook,times]{book}

=./wsuipa11
=./wsuipa10
=./wsuipa9
=./wsuipa8
=./wbxipa11
=./wbxipa10
=./wslipa10

\def\diatop[#1|#2]{{\setbox1=\hbox{{#1{}}}\setbox2=\hbox{{#2{}}}%
                    \dimen0=\ifdim\wd1>\wd2\wd1\else\wd2\fi%
                    \dimen1=\ht2\advance\dimen1by-1ex%
                    \setbox1=\hbox to1\dimen0{\hss#1\hss}%
                    \rlap{\raise1\dimen1\box1}%
                    \hbox to1\dimen0{\hss#2\hss}}}%
 
\def\ipa{\ipatwelverm}
 
\def\inva{\edef\next{\the\font}\ipa\char'000\next}%
\def\scripta{\edef\next{\the\font}\ipa\char'001\next}%
\def\nialpha{\edef\next{\the\font}\ipa\char'002\next}%
\def\invscripta{\edef\next{\the\font}\ipa\char'003\next}%
\def\invv{\edef\next{\the\font}\ipa\char'004\next}%
 
\def\crossb{\edef\next{\the\font}\ipa\char'005\next}%
\def\barb{\edef\next{\the\font}\ipa\char'006\next}%
\def\slashb{\edef\next{\the\font}\ipa\char'007\next}%
\def\hookb{\edef\next{\the\font}\ipa\char'010\next}%
\def\nibeta{\edef\next{\the\font}\ipa\char'011\next}%
 
\def\slashc{\edef\next{\the\font}\ipa\char'012\next}%
\def\curlyc{\edef\next{\the\font}\ipa\char'013\next}%
\def\clickc{\edef\next{\the\font}\ipa\char'014\next}%
 
\def\crossd{\edef\next{\the\font}\ipa\char'015\next}%
\def\bard{\edef\next{\the\font}\ipa\char'016\next}%
\def\slashd{\edef\next{\the\font}\ipa\char'017\next}%
\def\hookd{\edef\next{\the\font}\ipa\char'020\next}%
\def\taild{\edef\next{\the\font}\ipa\char'021\next}%
\def\dz{\edef\next{\the\font}\ipa\char'022\next}%
\def\eth{\edef\next{\the\font}\ipa\char'023\next}%
\def\scd{\edef\next{\the\font}\ipa\char'024\next}%
 
\def\schwa{\edef\next{\the\font}\ipa\char'025\next}%
\def\er{\edef\next{\the\font}\ipa\char'026\next}%
\def\reve{\edef\next{\the\font}\ipa\char'027\next}%
\def\niepsilon{\edef\next{\the\font}\ipa\char'030\next}%
\def\revepsilon{\edef\next{\the\font}\ipa\char'031\next}%
\def\hookrevepsilon{\edef\next{\the\font}\ipa\char'032\next}%
\def\closedrevepsilon{\edef\next{\the\font}\ipa\char'033\next}%
 
\def\scriptg{\edef\next{\the\font}\ipa\char'034\next}%
\def\hookg{\edef\next{\the\font}\ipa\char'035\next}%
\def\scg{\edef\next{\the\font}\ipa\char'036\next}%
\def\nigamma{\edef\next{\the\font}\ipa\char'037\next}
\def\ipagamma{\edef\next{\the\font}\ipa\char'040\next}%
\def\babygamma{\edef\next{\the\font}\ipa\char'041\next}%
 
\def\hv{\edef\next{\the\font}\ipa\char'042\next}%
\def\crossh{\edef\next{\the\font}\ipa\char'043\next}%
\def\hookh{\edef\next{\the\font}\ipa\char'044\next}%
\def\hookheng{\edef\next{\the\font}\ipa\char'045\next}%
\def\invh{\edef\next{\the\font}\ipa\char'046\next}%
 
\def\bari{\edef\next{\the\font}\ipa\char'047\next}%
\def\dlbari{\edef\next{\the\font}\ipa\char'050\next}% ``dotless bar i''
\def\niiota{\edef\next{\the\font}\ipa\char'051\next}%
\def\sci{\edef\next{\the\font}\ipa\char'052\next}%
\def\barsci{\edef\next{\the\font}\ipa\char'053\next}% ``barred small cap i''
 
\def\invf{\edef\next{\the\font}\ipa\char'054\next}%
 
\def\tildel{\edef\next{\the\font}\ipa\char'055\next}%
\def\barl{\edef\next{\the\font}\ipa\char'056\next}%
\def\latfric{\edef\next{\the\font}\ipa\char'057\next}%
\def\taill{\edef\next{\the\font}\ipa\char'060\next}%
\def\lz{\edef\next{\the\font}\ipa\char'061\next}%
\def\nilambda{\edef\next{\the\font}\ipa\char'062\next}%
\def\crossnilambda{\edef\next{\the\font}\ipa\char'063\next}%
 
\def\labdentalnas{\edef\next{\the\font}\ipa\char'064\next}%
\def\invm{\edef\next{\the\font}\ipa\char'065\next}%
\def\legm{\edef\next{\the\font}\ipa\char'066\next}%
 
\def\nj{\edef\next{\the\font}\ipa\char'067\next}%
\def\eng{\edef\next{\the\font}\ipa\char'070\next}%
\def\tailn{\edef\next{\the\font}\ipa\char'071\next}%
\def\scn{\edef\next{\the\font}\ipa\char'072\next}%
 
\def\clickb{\edef\next{\the\font}\ipa\char'073\next}%
\def\baro{\edef\next{\the\font}\ipa\char'074\next}%
\def\openo{\edef\next{\the\font}\ipa\char'075\next}%
\def\niomega{\edef\next{\the\font}\ipa\char'076\next}%
\def\closedniomega{\edef\next{\the\font}\ipa\char'077\next}%
\def\oo{\edef\next{\the\font}\ipa\char'100\next}%
 
\def\barp{\edef\next{\the\font}\ipa\char'101\next}%
\def\thorn{\edef\next{\the\font}\ipa\char'102\next}%
\def\niphi{\edef\next{\the\font}\ipa\char'103\next}%
 
\def\flapr{\edef\next{\the\font}\ipa\char'104\next}%
\def\legr{\edef\next{\the\font}\ipa\char'105\next}%
\def\tailr{\edef\next{\the\font}\ipa\char'106\next}%
\def\invr{\edef\next{\the\font}\ipa\char'107\next}%
\def\tailinvr{\edef\next{\the\font}\ipa\char'110\next}%
\def\invlegr{\edef\next{\the\font}\ipa\char'111\next}%
\def\scr{\edef\next{\the\font}\ipa\char'112\next}%
\def\invscr{\edef\next{\the\font}\ipa\char'113\next}%
 
\def\tails{\edef\next{\the\font}\ipa\char'114\next}%
\def\esh{\edef\next{\the\font}\ipa\char'115\next}%
\def\curlyesh{\edef\next{\the\font}\ipa\char'116\next}%
\def\nisigma{\edef\next{\the\font}\ipa\char'117\next}%
 
\def\tailt{\edef\next{\the\font}\ipa\char'120\next}%
\def\tesh{\edef\next{\the\font}\ipa\char'121\next}%
\def\clickt{\edef\next{\the\font}\ipa\char'122\next}%
\def\nitheta{\edef\next{\the\font}\ipa\char'123\next}%
 
\def\baru{\edef\next{\the\font}\ipa\char'124\next}%
\def\slashu{\edef\next{\the\font}\ipa\char'125\next}%
\def\niupsilon{\edef\next{\the\font}\ipa\char'126\next}%
\def\scu{\edef\next{\the\font}\ipa\char'127\next}%
\def\barscu{\edef\next{\the\font}\ipa\char'130\next}%
 
\def\scriptv{\edef\next{\the\font}\ipa\char'131\next}%
 
\def\invw{\edef\next{\the\font}\ipa\char'132\next}%
 
\def\nichi{\edef\next{\the\font}\ipa\char'133\next}%
 
\def\invy{\edef\next{\the\font}\ipa\char'134\next}%
\def\scy{\edef\next{\the\font}\ipa\char'135\next}%
 
\def\curlyz{\edef\next{\the\font}\ipa\char'136\next}%
\def\tailz{\edef\next{\the\font}\ipa\char'137\next}%
\def\yogh{\edef\next{\the\font}\ipa\char'140\next}%
\def\curlyyogh{\edef\next{\the\font}\ipa\char'141\next}%
 
\def\glotstop{\edef\next{\the\font}\ipa\char'142\next}%
\def\revglotstop{\edef\next{\the\font}\ipa\char'143\next}%
\def\invglotstop{\edef\next{\the\font}\ipa\char'144\next}%
\def\ejective{\edef\next{\the\font}\ipa\char'145\next}%
\def\reveject{\edef\next{\the\font}\ipa\char'146\next}%
 
\def\upt{\edef\next{\the\font}\ipa\char'154\next}%   These are IPA pointers
\def\downt{\edef\next{\the\font}\ipa\char'155\next}%
\def\leftt{\edef\next{\the\font}\ipa\char'156\next}%
\def\rightt{\edef\next{\the\font}\ipa\char'157\next}%
 
\def\upp{\edef\next{\the\font}\ipa\char'164\next}
\def\downp{\edef\next{\the\font}\ipa\char'165\next}%
\def\leftp{\edef\next{\the\font}\ipa\char'166\next}%
\def\rightp{\edef\next{\the\font}\ipa\char'167\next}%
 
\def\stress{\edef\next{\the\font}\ipa\char'150\next}%     primary stress
\def\secstress{\edef\next{\the\font}\ipa\char'151\next}%  secondary stress
 
\def\syllabic{\edef\next{\the\font}\ipa\char'152\next}%   syllabic marker
 
\def\corner{\edef\next{\the\font}\ipa\char'153\next}%
 
\def\halflength{\edef\next{\the\font}\ipa\char'160\next}
\def\length{\edef\next{\the\font}\ipa\char'161\next}
 
\def\underdots{\edef\next{\the\font}\ipa\char'162\next}%
 
\def\ain{\edef\next{\the\font}\ipa\char'163\next}
 
\def\overring{\edef\next{\the\font}\ipa\char'170\next}%
\def\underring{\edef\next{\the\font}\ipa\char'171\next}%
 
\def\open{\edef\next{\the\font}\ipa\char'172\next}%
 
\def\midtilde{\edef\next{\the\font}\ipa\char'173\next}%
\def\undertilde{\edef\next{\the\font}\ipa\char'174\next}%
 
\def\underwedge{\edef\next{\the\font}\ipa\char'175\next}%
 
\def\polishhook{\edef\next{\the\font}\ipa\char'176\next}%

\def\ipa {\ipatenrm}

\include{psfig}

\begin{document}
\date{}
\headers{Tzoukermann and Radev}{Use of Weighted Finite State Transducers}
\title{Use of Weighted Finite State Transducers in Part of Speech Tagging}
\author
{Evelyne Tzoukermann \\
Bell Labs, Lucent Technologies
\and
Dragomir R. Radev \\
Department of Computer Science, Columbia University 
}

\setcounter{chapter}{1}

\maketitle

\begin{abstract}
This paper addresses issues in part of speech disambiguation using
finite-state transducers and presents two main contributions to the
field. One of them is the use of finite-state machines for part of
speech tagging.  Linguistic and statistical information is represented
in terms of weights on transitions in {\em weighted\/} finite-state
transducers.  Another contribution is the successful combination of
techniques -- linguistic and statistical -- for word disambiguation,
compounded with the notion of word classes.
\end{abstract}
\section{Introduction}
 Finite-state machines have been extensively used in several areas of
natural language processing, including computational phonology,
morphology, and syntax.  Nevertheless, less has been done in the area
of part of speech disambiguation with finite-state transducers
\cite{Silberztein93,RocheSchabes95,ChanodTap95}.

Part of speech tagging consists of assigning to a word its
disambiguated part of speech in the sentential context in which this
word is used.  For languages which require morphological analysis, the
disambiguation is performed after the assignment of morphological
tags.  In this paper, we suggest two novel approaches for language
modeling for part of speech tagging.  The first is, in the absence of
sufficient training data, to use only word classes over lexical
probabilities.  This claim is well demonstrated and supported in
\cite{TzouRadGal95a,TzouRad96b}.  Second, we present a
complete system for part-of-speech disambiguation entirely implemented
within the framework of weighted finite-state transducers
\cite{PereiraRileySproat}.  Other works have been done using weighted
finite-state transducers (FST) with a combination of linguistic and
statistical techniques: \cite{SproatShihGaleChang} use weighted FSTs
to segment words in Chinese, and \cite{Sproat95} uses them for
multilingual text analysis.  The system we present disambiguates
unrestricted French texts with a success rate of over 96\%.

\section{System Overview}

\begin{figure}[htbp] 
\centerline{
\psfig{figure=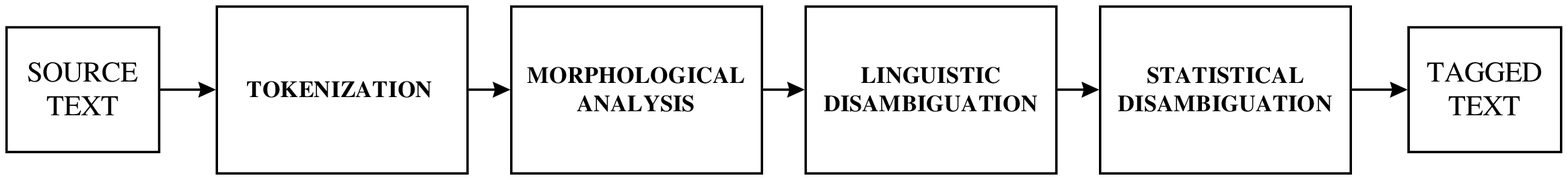,width=5in}
} 
\caption{System architecture.\label{fig:over}} 
\end{figure}

The input to the system is unrestricted French text; the unit over
which the algorithm functions is the sentence.  The system consists of
a cascade of FSTs, each of them corresponding to a different stage of
the disambiguation. The tagging process consists of several steps,
each involving the composition of the output of the previous stage
with one or more transducers.  Figure~\ref{fig:over} presents the main
stages of disambiguation.

\begin{enumerate}
\item {\bf Tokenization:} the input to the system is unprocessed
French text.  Each sentence is preprocessed according to several
criteria of normalization, such as treatment of compound conjunctions as
single units, treatment of uppercase words for proper names, and
acronyms.
\item {\bf Morphological analysis} is applied to the tokenized
sentence; see Table~\ref{example-sentence}, column 2.  We must point
out that there are over 250 tags for morphological analysis.  This
includes 45 verbal forms and 45 auxiliary forms, over 45 different
personal pronouns, etc.  These analyses were collapsed into 67 tags.
We use the larger tagset mostly at the negative constraint stage, as
it allows us to capture subtle agreement phenomena (see
Table~\ref{table-negative}). \footnote{Note that the word ``des'' in
Table~\ref{example-sentence} has three readings, namely (a) the
contraction of the preposition "de" and the article "les", (b) the
partitive article, (c) the indefinite article.  In the large tagset,
it is represented by three distinct tags; in the shorter tagset by
two tags only, i.e., the preposition tag for (a), and the article tag
for (b) and (c).}
\item {\bf Linguistic disambiguation:} the application of local
grammars expressing negative constraints, such as noun-pronoun non
agreement.
\item {\bf Statistical disambiguation:} n-gram probabilities are computed
on a training corpus and applied in terms of weights or costs on the FST
transitions. 
\end{enumerate}                                  
\begin{table}[htbp]
\caption{Morphologically tagged sentence.}     
%\begin{minipage}{\textwidth}
\vspace{.2cm}
\begin{tabular}{lll}
\hline\hline
{\bf Tokens}    & {\bf Full morphological analysis} & {\bf Tags}   \\\hline
le             & pron., {\bf def. masc. sg. art.}          & {\sc rdm}        \\
produit        & {\bf masc. sg. noun}, masc. sg. past part., 3rd pers. v. pres.& {\sc nms}  \\
liquide        & {\bf sg. adj.}, masc. sg. noun, 1st pers. v. ind./subj. pres., & \\
               & 2nd pers. v. imp,  3rd pers. v. ind./subj. pres. & {\sc jxs}        \\
qui            &  {\bf rel. pron.}, interr. pron.          		& {\sc br}   \\
entre          &  prep., 1st pers. v. ind./subj. pres., 2nd pers. v. imp., & \\
fem   bf                &  {\bf 3rd pers. v. ind.}/subj. {\bf pres.}  & {\sc 3spi}   \\
dans           &  masc. pl. noun, {\bf prep.}        		&	{\sc p}    \\
le             &  pron., {\bf def. masc. sg. art.}       		&	{\sc rdm}   \\
processus      &  {\bf masc. noun}        		 	&	{\sc nmx}   \\
des            &  {\bf prep.}, ind. pl. art., part. art.        &	{\sc p}   \\
photocopies    & {\bf fem. pl. noun}, 2nd pers. v. ind./subj. pres.  	&{\sc nfp} \\ \hline\hline
\end{tabular}                                    
%\end{minipage}
\label{example-sentence}
\end{table}                                      
The output text consists of the disambiguated French phrase,
see third column of Table~\ref{example-sentence} with
the corresponding analyses shown in bold in the second column.

\section{Weighted  for Morphological Analysis} \label{sec:morph}
\begin{figure}[htbp] 
\centerline{
\psfig{figure=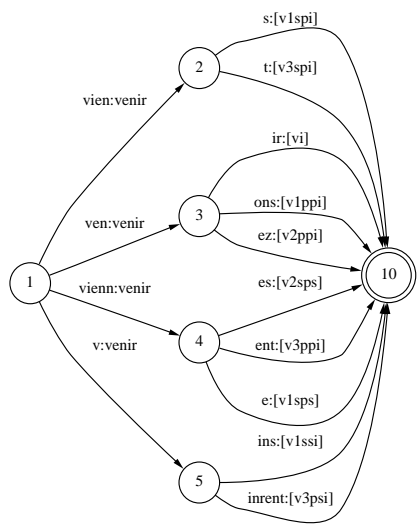,height=3in}
} 
\caption {FST showing some inflections of the verb ``venir'' (to come).}
\label{fig:vienn}
\end{figure}

The morphological transducer is developed within the framework of
finite-state morphology.  The system that we have developed goes from
lexical to surface form.  Phonological rules are applied separately to
compile verb, noun, and adjective stems.  For a given verb in French,
for example ``venir'' ({\em to come}), all the alternate base forms or
stems necessary for the complete verb inflection are computed before
the transduction from a French dictionary \cite{Boyer} and stored as
transitions in the list of arcs, thus forming the {\em arc-list}
dictionary \cite{TzoJacq97}.  This approach has been described in the
treatment of Spanish morphology \cite{TzouLib90}.
Figure~\ref{fig:vienn} shows the compiled base forms of the verb
``venir'' and some inflections associated with these stems.

The morphological FST is nondeterministic.  Weights are assigned to
the transitions of the FST.  The lower the weight, the more likely
that particular analysis will correspond to the proper disambiguation
of the word.  Thus, a word starting with an uppercase character will
have, as a proper noun, a higher weight than the same word if it
exists in the lexicon as a common noun.  For example, in the sentence
starting with ``March\'e conclu...'' ({\em completed (or done)
deal...})  the word ``March\'e'' is tagged {\sc npr} (proper noun),
{\sc nms} (masculine singular noun), and {\sc pp} (past participle).
In that context, it is more likely that ``March\'e'' is a common noun
rather than a proper one, thus the assignment of the lower cost to the
noun form.  Similarly, if a word contains only uppercase letters, it
can be tagged as an acronym, even though the acronym is not present in
the dictionary itself. In a similar fashion, the cost of tagging a
sequence of characters as an acronym is higher than the cost of
tagging the same sequence as a regular word.

\begin{figure}[htbp]
\centering
\input epsf
\epsfxsize=2.5in
\leavevmode
\epsffile{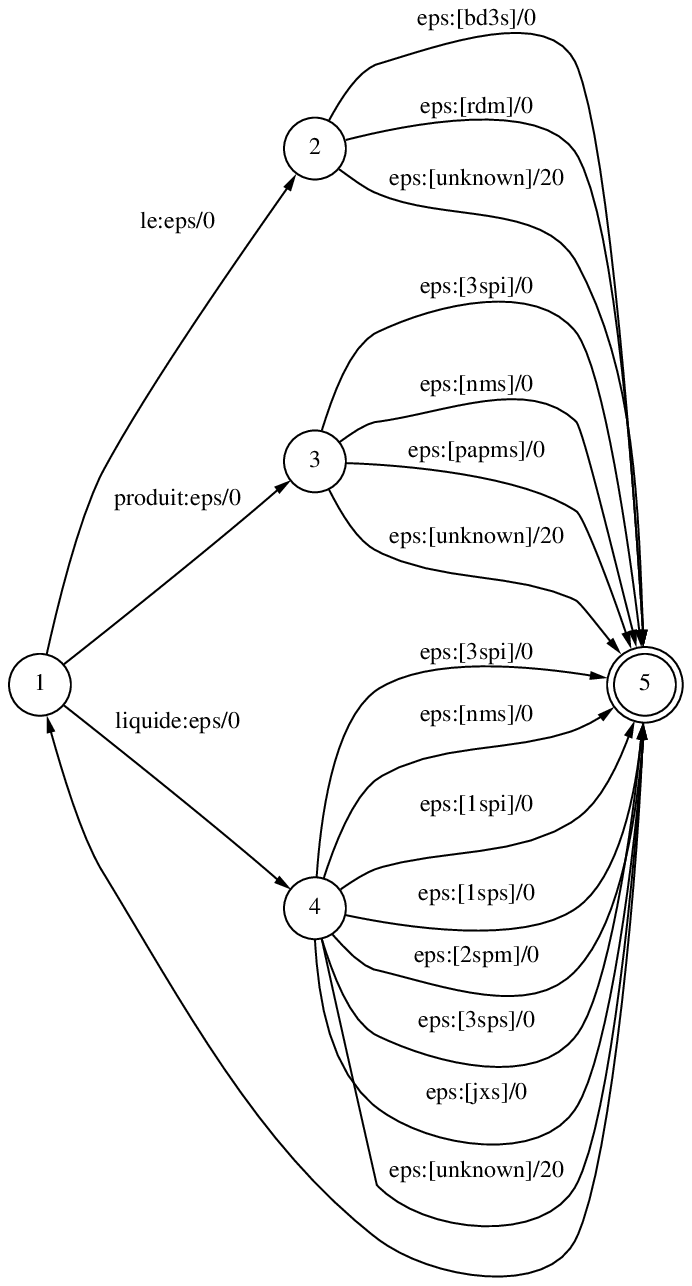}
\caption {Weighted sub-FST used to tag the input string ``le produit liquide''.}
\label{Figure:morph}
\end{figure}

Figure~\ref{Figure:morph} shows a finite-state automaton used to tag 
the sequence of three words ``le produit liquide''.  As an example, 
the word ``le'' and the morphological tags associated with it, namely 
{\sc [bd3s]} (3rd person singular direct pronoun), {\sc [rdm]} 
(masculine definite article), and {\sc [unknown]} are shown.  
At all stages of processing, we make sure that
composition of finite-state transducers doesn't fail. It happens that the
source text contains typos or grammatical errors. As a result, we always
allow for words to be tagged with the ``unknown'' tag (with a higher cost) 
in addition to their other tags. If at the end of processing, the
``unknown'' tag is the only tag remaining, the system will tag the
corresponding word as ``unknown''. If the ``unknown'' tag is not the only
one, it will have the highest cost of all and will not appear in the
output. 

\begin{figure}[htbp]
\centering
\input epsf
\epsfxsize=4.5in
\leavevmode
\epsffile{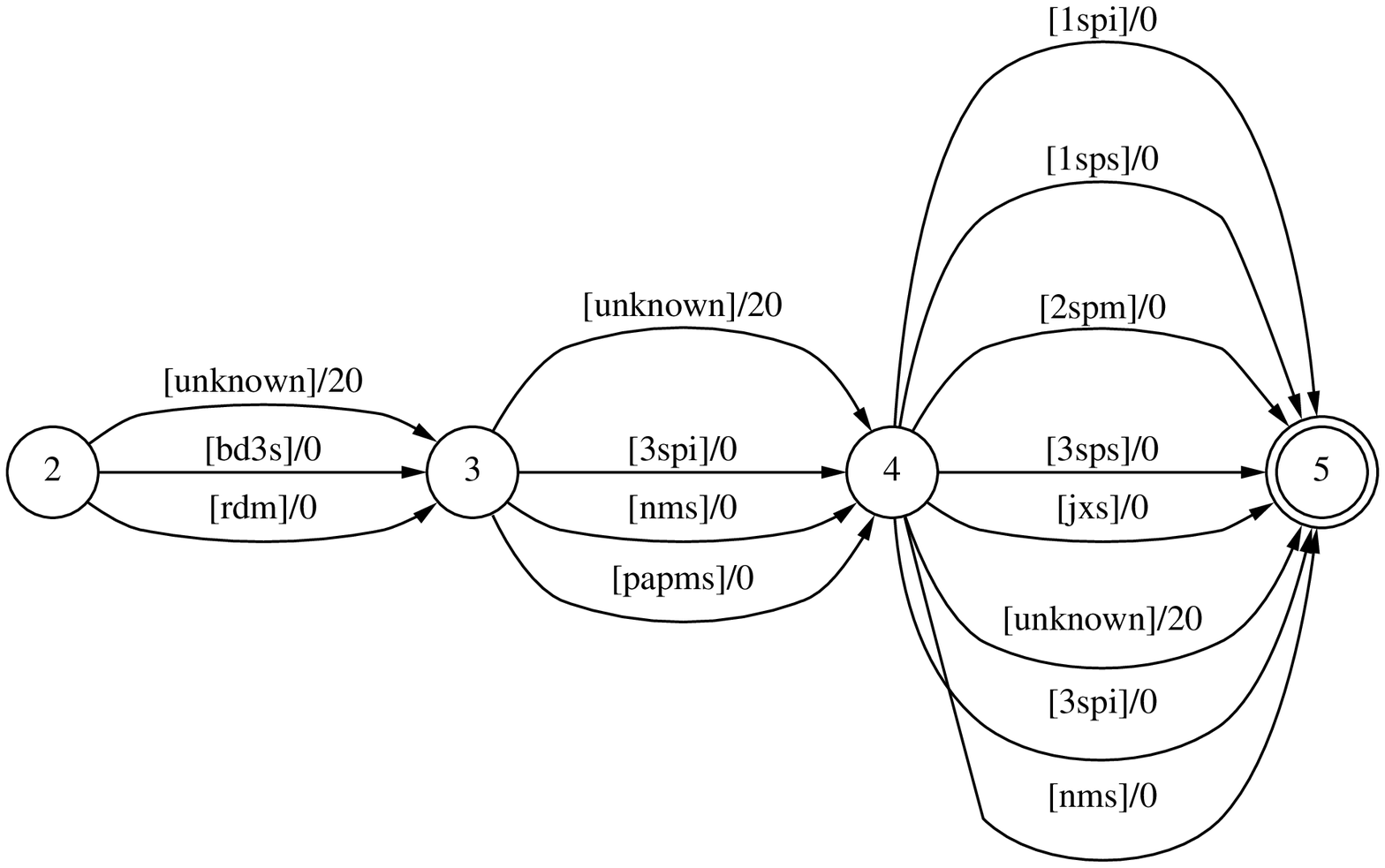}
\caption {Weighted FST representing the composition of the input string ``le produit liquide'' and the FST shown in Figure 3}
\label{Figt}
\end{figure}

Figure~\ref{Figt} shows the composition of the input string 
``le produit liquide'' and the FST shown in Figure~\ref{Figure:morph}.
One can clearly see the possible tags corresponding to the three
words in the input. As negative constraints and statistical rules
have not been applied yet, all weights are equal to $0$ except the ones
associated with the ``unknown'' tags.

In \cite{TzouRadGal96a}, we measured the ambiguity of French words in
unrestricted texts.  In comparing two corpora, one of about 100,000
tokens, the other of 200,000 tokens, we found out that 56\% of the
words are unambiguous, 27\% have two tags, 11\% have three tags, and
about 6\% have from four to eight tags.   The experiment showed three
important points: a) that over half of French words are ambiguous, b)
that their ambiguity varies from two tags for one fourth of the words
to eight tags for the other fourth of the words, and c) that the 
ambiguity is constant no matter the size of the corpus.

\subsection*{Training corpus and Genotypes}

Three separate corpora were used for training\footnote{ The corpora
consist of two different newspapers -- one corpus was extracted from
``Le Monde'' newspaper (corpus of the European Community Initiative,
1989, 1990), the other from the on-line collection of French news
distributed by the French Embassy in Washington D.C. between 1991 and
1994.}.  Their total size was of 76,162 manually tagged
tokens\footnote{We wish to thank Prof. Anne Abeill\'e and Thierry
Poibeau from the University of Paris for helping the manual tagging.}.
An additional corpus of 2,200 tokens was used for testing purposes.
The human tagger was given the output of the morphological analysis
and had to pick the proper tag from the set. 
 At the end of this time-consuming task, the total amount of
disambiguated text was still insufficient; lexical forms of words are
ignored and only their tags are considered.  Table~\ref{table-geno}
shows the distributions of genotypes in relation to tokens and word
types in the various corpora.  We use the term {\it genotype} to
capture the set of parts of speech a word can be tagged with.  For
example, the word ``liquide'' in Table~\ref{example-sentence} has a
genotype of {\sc [js nms v1s v2s v3s]}.  As shown in
Section~\ref{sec:stat}, probabilities are estimated on the genotypes
rather than the words (see \cite{TzouRad96b} for arguments on using
word class probabilities vs. lexical probabilities).  Genotypes play
an important role for smoothing probabilities.  By paying attention to
tags only and thus ignoring the words themselves, this approach
handles new words that have not been seen in the training corpus.  Our
approach is related to Cutting {\em et al.} \shortcite{CuttKupiec92},
who use the notion of word equivalence or ambiguity classes to
describe words belonging to the same part-of-speech categories.
However, they include only words under some frequencies of occurrence,
whereas our system uses word classes for every lexical item.  Notice
the ratio between the number of word types and the number of
genotypes.  In {\bf K1} for example, there are 219 genotypes for
10,006 tokens, whereas in {\bf K0}, 304 genotypes for 76,162 tokens,
i.e., only 38\% increase in the number of genotypes for a 661\% raise
in the corpus size.

\begin{table}[htbp]
\caption{Genotype distributions from the training corpora. }
\centering
%\begin{minipage}{\textwidth}
\vspace{.2cm}
\begin{tabular}{llll} 
\hline\hline
{\bf Corpora} & {\bf \# of tokens} & {\bf \# of types} & {\bf \# of
genotypes}\index{genotype} \\
\hline
{\bf K1} & 10006 & 2767 & 219   \\
{\bf K2} & 34636 & 4714 & 241 \\
{\bf K3} & 31520 & 5299 & 262 \\ 
{\bf K0 (K1-3)} & 76162 &10090 & 304 \\
\hline\hline
\end{tabular}
%\end{minipage}
\label{table-geno}
\end{table}

\section{Transducers of negative constraints}

Local grammars are used to represent linguistic information.  This
information is expressed in terms of negative constraints.  These local
grammars are somehow similar to the ones of \cite{Gross86,Mohri94,KarVouHeiAnt95},
and they reflect language generalities, allowing or disallowing
transitions from occurring.  For example, the most common example,
valid in several languages, states that an article ({\sc r}) cannot
precede a verb ({\sc v}) as shown in the constraint {\sc r v} in
Table~\ref{table-negative}.  This simple statement offers some
advantages: a) in the context of the two words ``le vol ({\em the
flight}), where ``le'' can be either an article ({\em the}) or a
personal pronoun ({\em it/him}), one can easily disambiguate ``le'';
if it precedes a noun (``vol''), it cannot be a pronoun, therefore it
is an article. b) in the context of the two words ``le manger'' ({\em
the nourishment} or {\em eat it}) where there is the additional
ambiguity of the word ``manger'' (noun or verb), instead of having
four readings, i.e. article-noun, article-verb, pronoun-noun,
pronoun-verb, two transitions are ruled out, namely article-verb and
pronoun-noun.  The two remaining readings will require an additional
word to disambiguate the tags in a trigram.
Table~\ref{table-negative} shows some examples of negative
constraints.  In order to favor local grammars over statistical
information, negative constraints have a cost lower than n-gram
genotypes obtained through statistics.
\begin{table}[htbp]
\caption{Sample negative constraints.}
%\begin{minipage}{\textwidth}
\vspace{.2cm}
\begin{tabular}{lll}
\hline\hline
{\bf Negative constraints}    & {\bf Parts of speech transitions}  \\ 
\hline
{\sc r v}      & article + verb \\
{\sc br1 v2}   & reflexive first person pronoun + second person verb \\
{\sc sb bd}    & sentence beginning + direct object personal pronoun \\
{\sc w j v}    & numeral + adjective + verb \\
{\sc rdm nfs}  & masculine definite article + feminine singular noun  \\ 
\hline\hline
\end{tabular}
%\end{minipage}
\label{table-negative}
\end{table}

All adjacencies that have to be ruled out by the tagger can be
expressed in such a way. The second rule in table~\ref{table-negative}
disallows the transition of a reflexive first person pronoun followed
by second person verb.  For instance, in the transition ``me vois''
({\em (I or you) see me}) where ``vois'' can be first or second
person, the first person is ruled out.  Agreement rules are
particularly well suited to be handled by this mechanism.  The last
transition in Table~\ref{table-negative} showed how a masculine
article cannot precede a feminine noun.  For example, the words ``le
mode'' ({\em the way} or {\em the fashion}) where ``mode'' can be
either masculine or feminine singular noun, the feminine form gets
ruled out to favor the masculine reading.

\begin{figure}[htbp] 
\centering
\input epsf
\epsfxsize=3in
\leavevmode
\epsffile{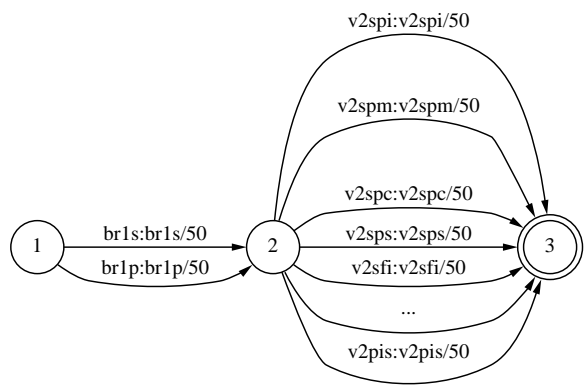}
\caption{Transducer of local grammars.} 
\label{fig:neg-cons}
\end{figure}

Stating negative rules in this manner offers an additional advantage
besides rule writing simplicity.  If the rule is generic for the tag,
only the generic representation will be written.  For instance, in the
first rule of Table~\ref{table-negative}, {\sc r} corresponds to all
the articles forms, which includes 13 tags, including {\sc rd}
(definite article), {\sc rdp} (definite partitive article), {\sc rdmp}
(definite masculine plural article), {\sc rdms} (definite masculine
singular article), etc.  If the rule focuses on gender agreement as is
the case in the last example of the table, it is possible to have a
more specific tag.  Figure \ref{fig:neg-cons} shows a transducer
corresponding to the local grammar {\sc br1 v2}.  In this particular
example {\sc br1} can be expanded into {\sc BR1P} (personal pronoun
reflexive 1st person plural) and {\sc BR1S} (personal pronoun reflexive
1st person singular), and {\sc v2} can be expanded into 30 tags,
including, among others {\sc V2PPI} (verb 2nd person plural present
indicative), {\sc V2SPM} (verb 2nd person singular present imperative),
{\sc V2SFI} (verb 2nd person singular future indicative), {\sc V2SIS}
(verb 2nd person singular imperfect subjunctive), all the second
person auxiliary forms, etc.  The negative constraint transducer is
used to increase the costs of certain paths in the automaton. When the
output of the morphological transducer is composed with the negative
constraint transducer, then the new transition costs are computed.
The result is that paths including transitions that correspond to
negative constraints will have an effective cost of infinity,
therefore will never be selected.  Since negative constraints are not
allowed to be violated, costs for "unknown" tags and negative
constraints were selected in such a way that paths including "unknown"
tags will have smaller costs than path with negative constraints.

A small number of constraints (in our case, only
77) can be expanded for all generic tags, thus creating a new set of
670 constraints.  This was achieved using a transducer compiling
rewriting rules that makes use of compositions of several transducers
\cite{MohriSproat96}.  This average expansion factor of 9 shows how
this rule writing mechanism can be economic for the linguist.

\section{Weighted FST for Statistical Tagging \label{sec:stat}}
We use n-grams of genotypes rather than word n-grams to estimate
frequencies.  Unigram, bigram, and trigram probabilities are computed
from the training corpus.  For example, bigram probabilities are
computed by estimating the sequence of two tags, $t_{i}$ and
$t_{i+1}$, given the two genotypes, $T_{i}$ and $T_{i+1}$, i.e.,
$P(t_{i},t_{i+1}|T_{i},T_{i+1})$, assuming that $t_{i} \in T_{i}$ and
$t_{i+1} \in T_{i+1}$.  For all parts of speech, the weights are
derived from the frequency of a given genotype in context within the
training corpus.  Weights are associated with each n-gram and applied
during tagging.  Due to their distribution and to the disambiguation
process, some words such as proper nouns, acronyms, and unknown words,
are assigned higher weights.

\begin{figure}[htbp]
\centering
\input epsf
\epsfxsize=4.5in
\leavevmode
\epsffile{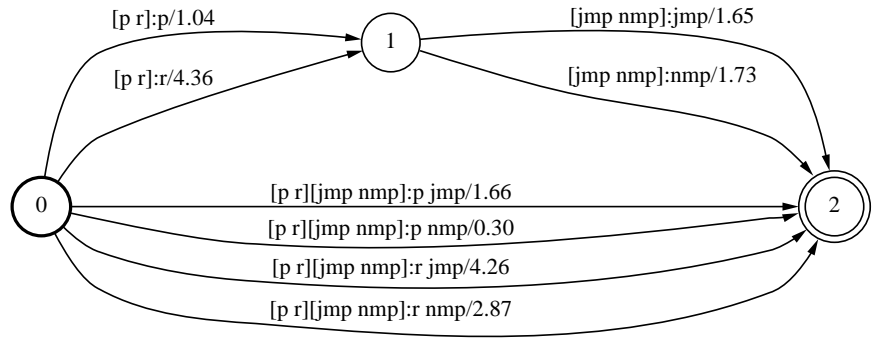}
\caption {Example of a Weighted FST which tags the genotype bigram {\sc [p r] [jmp nmp]} \label{figure-statist}}
\end{figure}

Figure~\ref{figure-statist} presents a bigram genotype showing all the
transitions and weights, and Table~\ref{table-statist} demonstrates how
weights are computed for a specific bigram and how these weights are
used to make a tagging decision.  The bigram {\sc [p r] [jmp nmp]}
occurs 141 times in the training corpus, and corresponds to the
possible word ``des'' ({\em the, of the}) which has for genotype {\sc [p
r]} (preposition, article) and the possible word ``bons'' ({\em good
ones, good}) with the genotype {\sc [jmp nmp]} (masculine plural
adjective, masculine plural noun).  The bigram is generated
automatically from the training corpus; observe in
Figure~\ref{figure-statist} that there are 8 possible readings for the
bigram (4 unigram combinations and 4 bigrams).  On the one hand, the
four combinations of the separate unigrams going from state 0 to 1 and
from 1 to 2, each one appearing in the training corpus.  In these
cases, the final weights correspond to the sum of the values of {\sc
[p]} and {\sc [jmp]}, i.e. 1.66, {\sc [p]} and {\sc [nmp]} with a
weight of 0.30, {\sc [r]} and {\sc [jmp]} with a weight of 4.26, and
{\sc [r]} and {\sc [nmp]} with a weight of 2.87.  On the other hand,
the sub-FST that corresponds to this bigram of genotypes will have
[{\sc p r] [jmp nmp]} on its input and all 4 possible taggings on its
output, as illustrated in Table~\ref{table-statist}.  Each tagging
sequence has a different weight. Assume that $f$ is the sum of all weights in a
genotype bigram and $f_t$ is the number of cases where 
$t$ occurs. For all possible taggings $t$ (in this example there are 4
possible taggings), the weight of the transition for tagging $t$ is
the negative logarithm of $f_t$ divided by $f$: $- \log(f_t/f)$.  Thus, the
decision {\sc p jmp} appears with the weight 1.66, the decision {\sc p
nmp} with the weight 0.30, the decision {\sc r jmp} with the weight
4.26, and finally the decision {\sc r nmp} with the weight 2.87.  Out
of these eight combinations, the lowest cost is 0.30, which means that
the bigram {\sc p nmp} will be selected. 
%DRAGO  Perhaps we could say how the example is appropriate to 
% answer the reviewer.
\begin{table}[htbp]
\caption{An example of cost computation for the bigram FST {\sc [p r] [jmp nmp]}.}
\centering
%\begin{minipage}{\textwidth}
\vspace{.2cm}
\begin{tabular}{llrr} 
\hline\hline
{\bf genotype bigram} & {\bf tagging} & {\bf frequency} & {\bf weight} \\
\hline
{\sc [p r] [jmp nmp]} & {\sc p, jmp} & 27/141 & 1.66 \\
                & {\bf {\sc p, nmp}} & {\bf 104/141} & {\bf 0.30} \\
                & {\sc r, jmp} & 2/141 & 4.26 \\
                & {\sc r, nmp} & 8/141 & 2.87 \\
\hline\hline
\end{tabular}
%\end{minipage}
\label{table-statist}
\end {table}

\begin{figure}[htbp]
\centering
\input epsf
\epsfxsize=4.5in
\leavevmode
\epsffile{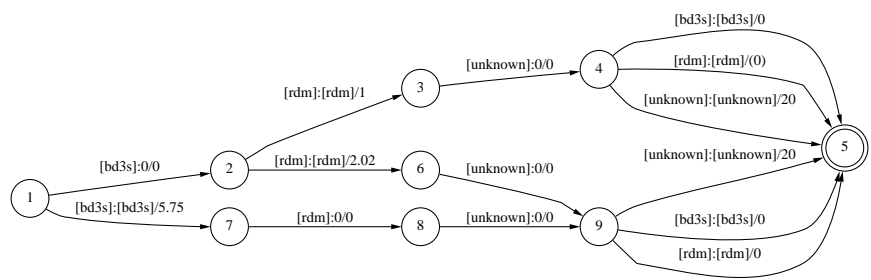}
\caption {Weighted FST representing the genotype unigram {\sc [bd3s rdm]} corresponding to the word ``le'' in the sample sentence.} 
\label{Sent1.bp}
\end{figure}

\section{Contextual probabilities via bigram and trigram genotypes
\label{cont-prob}}

Using genotypes at the unigram level tends to result in
overgeneralization, due to the fact that the genotype sets are too
coarse.  In order to increase the accuracy of part-of-speech disambiguation,
we need to give priority to trigrams over bigrams, and to bigrams over
unigrams.

In a way similar to decision trees, Table~\ref{ngram} shows how the
use of context allows for better disambiguation of genotype.  We have
considered a typical ambiguous genotype {\sc [jmp nmp]}, corresponding
to a word such as ``petits'' (small) which can be either masculine
plural adjective (small) or masculine plural noun (small ones), which
occurs 607 times in the training corpus, almost evenly distributed
between the two alternative tags, {\sc jmp} and {\sc nmp}.  As a
result, if only unigram training data is used, the best candidate for
that genotype would be {\sc jmp}, occurring 316 out of 607 times.
However, choosing {\sc jmp} only gives us 52.06\% accuracy.
Table~\ref{ngram} clearly demonstrates that the contextual information
around the genotype will bring this percentage up significantly.  As
an example, let us consider the 5th line of Table~\ref{ngram}, where
the number 17 is marked with a square.  In this case, we know that the
{\sc [jmp nmp]} genotype has a right context consisting of the
genotype [p r] (4th column, 5th line).  In this case, it is no longer
true that {\sc jmp} is the best candidate.  Instead, {\sc nmp} occurs
71 out of 91 times and becomes the best candidate.  Overall, for all
possible left and right contexts of {\sc [jmp nmp]}, the guess based
on both the genotype and the single left or right contexts will be
correct 433 times out of 536 (or 80.78\%).  In a similar fashion, the
three possible trigram patterns (Left, Middle, and Right) are shown in
lines 18-27.  They show that the performance based on trigrams is
95.90\%.  Disambiguation results are provided in Table~\ref{dis-res}.
This particular example provides strong evidence of the
usefulness of contextual disambiguation with genotypes.  The fact that
this genotype, very ambiguous as a unigram (52.06\%), can be
disambiguated as a noun or adjective according to context at the
trigram stage with 95.90\% accuracy demonstrates the strength of our
approach.

\begin{table}[htbp]
\caption{Influence of context for n-gram genotype disambiguation. \label{ngram}}
\centering
\scriptsize
\vspace{.2cm}
\begin{tabular}{llrllrrrrrr} \hline \hline
{\bf n-gram} & {\bf pos.} &{\bf total} & {\bf genotype} & {\bf
decision} & {\bf distr.} & {\bf correct} & {\bf total} \\ 
\hline
Unigram &    & 607 & {\bf [jmp nmp]}       & {\bf jmp}    & {\bf 316} &
316 & 607 \\ 
\cline{5-6}
  &    &     &                 & {\bf nmp}    & 291 &     &      \\ \hline
Bigram & Left  & 230 & {\bf [jmp nmp]}[x]    & {\bf jmp}, x & {\bf 71} & 71  & 102 \\ 
\cline{5-6}
  &    &     &                 & {\bf nmp}, x & 31  &     &      \\ 
\cline{4-8}
  &    &     & {\bf [jmp nmp]}[p r]  & {\bf jmp}, p & \fbox{{\bf 17}}  & 71  & 91   \\ 
\cline{5-6}
  &    &     &                 & {\bf jmp}, r & 3   &     &      \\ 
\cline{5-6}
  &    &     &                 & {\bf nmp}, p & 71  &     &      \\ 
\cline{4-8}
  &    &     & {\bf [jmp nmp]}[nmp]  & {\bf jmp}, nmp & {\bf 23} & 23 & 24    \\ 
\cline{5-6}
  &    &     &                 & {\bf nmp}, nmp & 1  &    &      \\ 
\cline{4-8}
  &    &     & {\bf [jmp nmp]}[a]    & {\bf jmp}, a & {\bf 13}  & 13  & 13  \\ 
\cline{2-8}
  & Right  & 306 & [p r]{\bf [jmp nmp]}  & p, {\bf jmp} & 27  & 112 & 141 \\ 
\cline{5-6}
  &    &     &                 & p, {\bf nmp} & {\bf 104} &     &     \\ 
\cline{5-6}
  &    &     &                 & r, {\bf jmp} & 2   &     &           \\ 
\cline{5-6}
  &    &     &                 & r, {\bf nmp} & {\bf 8}   &     &     \\ 
\cline{4-8}
  &    &     & [b r]{\bf [jmp nmp]}  & r, {\bf jmp} & 22 & 72   & 94  \\ 
\cline{5-6}
  &    &     &                 & r, {\bf nmp} & {\bf 72}  &     &     \\ 
\cline{4-8}
  &    &     & [nmp]{\bf [jmp nmp]}      & nmp, {\bf jmp}    &{\bf 71} & 71 & 71   \\ \hline
Trigram & Left  & 32  & {\bf [jmp nmp]}[p r][nms]   & {\bf nmp}, p, nms & {\bf 21} & 21 & 21 \\ 
\cline{4-8}
  &    &     & {\bf [jmp nmp]}[jmp nmp][x] & {\bf jmp}, jmp, x & 3  & 8  & 11    \\ 
\cline{5-6}
  &    &     &                       & {\bf nmp}, jmp, x & {\bf 8}  &   &        \\ 
\cline{2-8}
  & Middle  & 44  & [p r]{\bf [jmp nmp]}[p r] & p, {\bf nmp}, p & {\bf 23}  & 23 & 23 \\ 
\cline{4-8}
  &    &     & [b r]{\bf [jmp nmp]}[p r] & r, {\bf nmp}, p & {\bf 19}  & 19 & 21   \\ 
\cline{5-6}
  &    &     &                       & r, {\bf jmp}, p & 2  &    &       \\ 
\cline{2-8}
  & Right  & 46  & [p r][nmp]{\bf [jmp nmp]} & p, nmp, {\bf jmp} & {\bf 27}  & 29 & 29 \\ 
\cline{5-6}
  &    &     &                     & r, nmp, {\bf jmp} & {\bf 2}  &   &     \\ 
\cline{4-8}
  &    &     & [n z][p r]{\bf [jmp nmp]} & z, p, {\bf nmp} & {\bf 16} & 17 & 17     \\ 
\cline{5-6}
  &    &     &                     & z, r, {\bf nmp} & {\bf 1}  &    &     \\ \hline \hline
\end{tabular}
\end{table}

\begin{table}[htbp]
\caption{Evaluation of the predictive power of contextual genotypes.}
\label{dis-res}
\centering
\vspace{.2cm}
\begin{tabular}{lrrr} \hline \hline
{\bf n-gram} & {\bf cor.} & {\bf total} & {\bf accuracy} \\ \hline
Unigram      & 316        & 607         & 52.06\%        \\
Bigram       & 433        & 536         & 80.78\%        \\
Trigram      & 117        & 122         & 95.90\%        \\ \hline \hline
\end{tabular}
\label{ngram2}
\end{table}

\subsection*{Smoothing probabilities with genotypes}
In the context of a small training corpus, the problem of sparse data
is more serious than with a larger tagged corpus.  Genotypes play an
important role for smoothing probabilities.  By paying attention to
tags only and thus ignoring the words themselves, this approach handles new
words that have not been seen in the training corpus.
Table~\ref{gen-occ} shows how the training corpus provides coverage
for n-gram genotypes that appear in the test corpus.
It is interesting to notice that only 12 out of 1564 unigram genotypes
(0.8\%) are not covered.  The training corpus covers 71.4\% of the
bigram genotypes that appear in the test corpus and 22.2\% of the
trigrams.

\begin{table}[htbp]
\caption{Coverage in the training corpus of n-gram genotypes that
appear in the test corpus. }
\centering
\vspace{.2cm}
\begin{tabular}{lccc}\hline \hline
     & {\bf test corpus}   & {\bf training  corpus}  & \\ 
     & {\bf \# of genotypes}   & {\bf \# of genotypes} & accuracy \\ \hline
1-grams  & 1564   &  1552   &  (99.2 \%)\\
2-grams  & 1563   &  1116   &  (71.4 \%)\\
3-grams  & 1562   &   346   &  (22.2 \%) \\ \hline \hline
\end{tabular}
\label{gen-occ}
\end{table}

\section{Related Research}
Approaches to part of speech taggers can be divided into two types:
Markov-model based taggers on the one hand
\cite{BahlMercer76,LeeGarAtw83,Merialdo94,DeRose88,Church89b,CuttKupiec92},
and rule-based part of speech taggers
\cite{KleinSimm63,Brill92,Voutilai93} on the other.  Even though there has
been a 
recent surge of interest in the application of finite-state automata
to NLP issues, work has only started in part of speech tagging.  Roche
and Schabes (1995) present a part-of-speech tagger based on
finite-state transducers; they use Brill's part of speech tagger and
convert the rules into finite-state transducers.  Operations are
accomplished on the transducers, such as the application of a Local
Extension function.  Transducers are converted into subsequential
ones, to be deterministic.  The goal of the operation is to optimize
the system in terms of time and execution speed, which is crucial for
a working system.  The work does not focus on the disambiguation per
se, but rather, on the conversion of transducers into deterministic
subsequential ones.

Chanod and Tapanainen \shortcite{ChanodTap95,ChanodTap95b} compare two
frameworks for tagging French, a statistical one, based on the Xerox
tagger \cite{CuttKupiec92}, and another based on linguistic
constraints only.  The constraint-based tagger is proven to have better
performance than the statistical one, since rule writing is easier to
handle and to control than adjusting the parameters of the statistical
tagger.  It is difficult to compare any kind of performance with ours
since their tagset is very small, i.e. 37 tags (compared to our two
tagsets of 67 and 253 tags), including a number of word-specific tags
which further reduces the number of tags, and does not account for
several morphological features, such as gender, number for pronouns,
etc.  To be properly done, the comparison would involve major changes
in our system since local grammars could not be applied as is, and
n-gram statistics should be re-computed.
Moreover, categories that
can be very ambiguous, such as coordinating conjunctions,
subordinating conjunctions, relative and interrogative pronouns tend
to be collapsed; consequently, the disambiguation is simplified and
it is not straightforward to compare results.

\section{Results and Conclusion}
Using weighted FSTs to couple statistic and linguistic information
has shown to be highly successful in part of speech tagging. 
The size of the different modules of the system is presented in Table~\ref{tab:res}:
\begin{table}[htbp]
\caption{Size of the different transducers.} 
\centering
\vspace{.2cm}
\begin{tabular}{cccc}        \hline \hline
& {\bf  Morphology}  &  {\bf Negative constraints} & {\bf Ngram genotypes} \\ \hline
Number of states & 810,263 & 181 & 12,718 \\ 
Number of arcs   & 914,561 & 39,549 & 2,520,846 \\ \hline \hline
\end{tabular}
\label{tab:res}
\end{table}
Our system correctly disambiguates 96\% of words in unrestricted texts. 
We ran an experiment using 10,000 words of training corpus in order to
measure the improvement of n-gram disambiguation.  We tested our tagger on
a 1,000-word corpus.
Table~\ref{table:res} shows how the performance of the tagger improves from 
92.1\% using only unigrams to 96.0\% using unigrams, bigrams, trigrams, and 
negative constraints.

\begin{table}[htbp]
\caption{Tagger performance with {\it n}-gram probabilities and
negative constraints.} 
\centering
\vspace{.2cm}
\begin{tabular}{cccc}        \hline \hline
& {\bf  1-grams}  &  {\bf 1, 2 -grams } & {\bf neg. cons and 1, 2, 3 -grams} \\ \hline
10K-word corpus & 92.1\% & 93.4\% & 96.0\% \\ \hline \hline
\end{tabular}
\label{table:res}
\end{table}

We demonstrated that, in the absence of more training data, the use of
genotypes captures linguistic generalities about words.  Additionally,
genotypes are used for smoothing which seriously reduces the problem
of sparse data.  Bigram and trigram genotypes capture the pattern of
tags in context.  The system has been used in automatic indexing
applications and text-to-speech system for French.  In text-to-speech,
words having the same orthography and a different pronunciation, can
be identified via their part-of-speech.  This is the case of verb/noun
category where words like ``pr\'esident'' can be pronounced
either [presid\~\scripta] (when it is a noun) or
[presid(\schwa)] (when it is a verb), the noun/verb words such as
``est'' [\niepsilon st] (noun) and [\niepsilon] (verb).  Knowing
parts of speech for text-to-speech applications also permits to
compute better intonational contours.  We are planning to utilize
additional FST tools for local grammars so that shallow syntactic
units can be studied and analyzed.

%\bibliography{/usr/evelyne/paper/Biblio/refs}
%\bibliography{refs}

\begin{thebibliography}{}

\bibitem[\protect\citename{Bahl and Mercer}1976]{BahlMercer76}
Lalit~R. Bahl and Robert~L. Mercer.
\newblock 1976.
\newblock Part-of-speech assignment by a statistical decision algorithm.
\newblock {\em IEEE International Symposium on Information Theory}, pages
  88--89.

\bibitem[\protect\citename{Boyer}1993]{Boyer}
Martin Boyer.
\newblock 1993.
\newblock {\em Dictionnaire du fran\protect{\c{c}}ais}.
\newblock Hydro-Qu\protect{\'{e}}bec, GNU General Public License, Qu\protect{\'{e}}bec,
  Canada.

\bibitem[\protect\citename{Brill}1992]{Brill92}
Eric Brill.
\newblock 1992.
\newblock A simple rule-based part of speech tagger.
\newblock In {\em Third Conference on Applied Computational Linguistics},
  Trento, Italy.

\bibitem[\protect\citename{Chanod and Tapanainen}1995]{ChanodTap95}
Jean-Pierre Chanod and Pasi Tapanainen.
\newblock 1995.
\newblock Tagging \protect{F}rench -- comparing a statistical and a
  constraint-based method.
\newblock In {\em EACL}, Dublin, Ireland. Association for
  Computational Linguistics - European Chapter.

\bibitem[\protect\citename{Chanod and Tapanainen}1995a]{ChanodTap95b}
Jean-Pierre Chanod and Pasi Tapanainen.
\newblock 1995a.
\newblock Creating a tagset, lexicon and guesser for a \protect{F}rench tagger.
\newblock In {\em EACL SIGDAT Workshop}, Dublin, Ireland. Association for
  Computational Linguistics - European Chapter.

\bibitem[\protect\citename{Church}1989]{Church89b}
Kenneth~W. Church.
\newblock 1989.
\newblock A stochastic parts program noun phrase parser for unrestricted text.
\newblock In {\em IEEE Proceedings of the ICASSP}, pages 695--698, Glasgow.

\bibitem[\protect\citename{Cutting \bgroup et al.\egroup }1992]{CuttKupiec92}
Doug Cutting, Julian Kupiec, Jan Peterson, and Penelope Sibun.
\newblock 1992.
\newblock A practical part-of-speech tagger.
\newblock Trento, Italy. Proceedings of the Third Conference on Applied Natural
  Language Processing.

\bibitem[\protect\citename{DeRose}1988]{DeRose88}
Stephen DeRose.
\newblock 1988.
\newblock Grammatical category disambiguation by statistical optimization.
\newblock {\em Computational Linguistics}, 14(1):31--39.

\bibitem[\protect\citename{Gross}1986]{Gross86}
Maurice Gross.
\newblock 1986.
\newblock {\em Grammaire transformationnelle du fran\protect{\c{c}}ais - 1.
  Syntaxe du verbe}.
\newblock Cantil\`ene, 92240 Malakoff.

\bibitem[\protect\citename{Kaplan and Kay}1994]{kk}
Ronald~M. Kaplan and Martin Kay.
\newblock 1994.
\newblock Regular models of phonological rule systems.
\newblock {\em Computational Linguistics}, 20(3).

\bibitem[\protect\citename{Karlsson \bgroup et al.\egroup
  }1995]{KarVouHeiAnt95}
Fred Karlsson, Atro Voutilainen, Juha {Heikkil\"{a}}, and Atro Antilla.
\newblock 1995.
\newblock {\em Constraint Grammar: A Language-Independent System for Parsing
  Unrestricted Text}.
\newblock Mouton de Gruyter, Berlin, New York.

\bibitem[\protect\citename{Karttunen}1983]{Karttunen83a}
Lauri Karttunen.
\newblock 1983.
\newblock Kimmo: A general morphological processor.
\newblock In {\em Texas Linguistic Forum}, volume~22, pages 165--186.

\bibitem[\protect\citename{Klein and Simmons}1963]{KleinSimm63}
S.~Klein and R.~F. Simmons.
\newblock 1963.
\newblock A grammatical approach to grammatical tagging coding of
  \protect{E}nglish words.
\newblock {\em JACM}, 10:334--347.

\bibitem[\protect\citename{Koskenniemi}1983]{Koskenniemi83}
Kimmo Koskenniemi.
\newblock 1983.
\newblock {\em Two-Level Morphology: a General Computational Model for
  Word-Form Recognition and Production}.
\newblock {Ph.D.} thesis, University of Helsinki, Helsinki.

\bibitem[\protect\citename{Leech \bgroup et al.\egroup }1983]{LeeGarAtw83}
Geoffrey Leech, Roger Garside, and Erik Atwell.
\newblock 1983.
\newblock Automatic grammatical tagging of the {LOB} corpus.
\newblock {\em ICAME News}, 7:13--33.

\bibitem[\protect\citename{Merialdo}1994]{Merialdo94}
Bernard Merialdo.
\newblock 1994.
\newblock Tagging {E}nglish text with a probabilistic model.
\newblock {\em Computational Linguistics}, 20(2):155--172.

\bibitem[\protect\citename{Mohri and Sproat}1996]{MohriSproat96}
Mehryar Mohri and Richard Sproat.
\newblock 1996.
\newblock An efficient compiler for weighted rewrite rules.
\newblock In {\em 34th Annual Meeting of the Association for Computational
  Linguistics}, pages 231--238, Santa Cruz, Ca. Association for Computational
  Linguistics.

\bibitem[\protect\citename{Mohri}1994]{Mohri94}
Mehryar Mohri.
\newblock 1994.
\newblock Syntactic analysis by local grammars automata: an efficient
  algorithm.
\newblock In {\em COMPLEX '94}, Budapest, Hungary. Proceedings of the
  International Conference on Computational Lexicography.

\bibitem[\protect\citename{Pereira \bgroup et al.\egroup
  }1994]{PereiraRileySproat}
Fernando Pereira, Michael Riley, and Richard Sproat.
\newblock 1994.
\newblock Weighted rational transductions and their application to human
  language processing.
\newblock In {\em {ARPA} Workshop on Human Language Technology}, pages
  249--254. Advanced Research Projects Agency, March 8--11.

\bibitem[\protect\citename{Roche and Schabes}1995]{RocheSchabes95}
Emmanuel Roche and Yves Schabes.
\newblock 1995.
\newblock Deterministic part-of-speech tagging with finite-state transducers.
\newblock {\em Computational Linguistics}, 21(2).

\bibitem[\protect\citename{Roche}1993]{Roche93}
Emmanuel Roche.
\newblock 1993.
\newblock {\em Analyse syntaxique transformationnelle du fran\protect{\c{c}}ais
  par transducteur et lexique-grammaire}.
\newblock {Ph.D.} thesis, Universit\protect{\'{e}} Paris 7, Paris, France.

\bibitem[\protect\citename{Silberztein}1993]{Silberztein93}
Max Silberztein.
\newblock 1993.
\newblock {\em Dictionnaires \'electroniques et analyse automatique de textes:
  le syst\`eme INTEX}.
\newblock Masson, Paris, France.

\bibitem[\protect\citename{Sproat \bgroup et al.\egroup
  }1996]{SproatShihGaleChang}
Richard Sproat, Chilin Shih, William Gale, and Nancy Chang.
\newblock 1996.
\newblock A stochastic finite-state word-segmentation algorithm for chinese.
\newblock {\em Computational Linguistics}, 22(3).

\bibitem[\protect\citename{Tapanainen and Voutilainen}1993]{TapVout93}
Pasi Tapanainen and Atro Voutilainen.
\newblock 1993.
\newblock Ambiguity resolution in a reductionistic parser.
\newblock In {\em Association for Computational Linguistics - European
  Chapter}, pages 394--403, Utrecht, Netherlands.

\bibitem[\protect\citename{Sproat}1995]{Sproat95}
Richard Sproat.
\newblock 1995.
\newblock A finite-state architecture for tokenization and grapheme-to-phoneme
  conversion in multilingual text analysis.
\newblock In {\em Proceedings of the ACL SIGDAT Workshop, Dublin, Ireland}.
  ACL.

\bibitem[\protect\citename{Tzoukermann and Jacquemin}1997 to appear]{TzoJacq97}
Evelyne Tzoukermann and Christian Jacquemin.
\newblock 1997, to appear.
\newblock Analyse automatique de la morphologie d\protect{\'{e}}rivationnelle.
\newblock Lille, France. Mots possibles et mots existants.

\bibitem[\protect\citename{Tzoukermann and Liberman}1990]{TzouLib90}
Evelyne Tzoukermann and Mark~Y. Liberman.
\newblock 1990.
\newblock A finite-state morphological processor for \protect{S}panish.
\newblock In {\em Proceedings of the Thirteenth International Conference on
  Computational Linguistics}, Helsinki, Finland. International Conference on
  Computational Linguistics.

\bibitem[\protect\citename{Tzoukermann \bgroup et al.\egroup }1997 to
  appear]{TzouRadGal96a}
Evelyne Tzoukermann, Dragomir~R. Radev, and William~A. Gale, 1997, to appear.
\newblock {\em Tagging French Without Lexical Probabilities}.
\newblock Kluwer.

\bibitem[\protect\citename{Tzoukermann and Radev}1996]{TzouRad96b}
Evelyne Tzoukermann and Dragomir~R. Radev.
\newblock 1996.
\newblock Using word class for part-of-speech disambiguation.
\newblock In {\em Fourth Workshop on Very Large Corpora}, pages 1--13,
  Copenhagen, Denmark. International Conference on Computational Linguistics.

\bibitem[\protect\citename{Tzoukermann \bgroup et al.\egroup
  }1995]{TzouRadGal95a}
Evelyne Tzoukermann, Dragomir~R. Radev, and William~A. Gale.
\newblock 1995.
\newblock Combining linguistic knowledge and statistical learning in
  \protect{F}rench part-of-speech tagging.
\newblock In {\em EACL SIGDAT Workshop}, pages 51--57, Dublin, Ireland.
  Association for Computational Linguistics - European Chapter.

\bibitem[\protect\citename{Voutilainen}1993]{Voutilai93}
Atro Voutilainen.
\newblock 1993.
\newblock \protect{NP}tool, a detector of \protect{E}nglish noun phrases.
\newblock Columbus, Ohio. Proceedings of the Workshop on very large corpora.

\end{thebibliography}

\end{document}